\documentclass{jors}

\usepackage[numbers]{natbib}
\definecolor{mygray}{gray}{0.6}

\begin{document}

{\bf The BioDynaMo Project} \\

\rule{\textwidth}{1pt}

\section*{Paper Authors}


{1. Bauer, Roman; \\
2. Breitwieser, Lukas; \\
3. Di Meglio, Alberto; \\
4. Johard, Leonard; \\
5. Kaiser, Marcus; \\
6. Manca, Marco;\\
7. Mazzara, Manuel \\
8. Talanov, Max}

\section*{Paper Author Roles and Affiliations}
{1. Newcastle University, United Kingdom, email: roman.bauer@newcastle.ac.uk \\
2. CERN openlab, Switzerland, email: lukas.johannes.breitwieser@cern.ch \\
3. CERN openlab, Switzerland, email: alberto.di.meglio@cern.ch \\
4. Innopolis University, Russian Federation, email: l.johard@innopolis.ru \\
5. Newcastle University, United Kingdom, email: marcus.kaiser@newcastle.ac.uk \\
6. CERN openlab, Switzerland, email: marco.manca@cern.ch \\
7. Innopolis University, Russian Federation, email: m.mazzara@innopolis.ru \\
8. Kazan Federal University, email: max.talanov@gmail.com
}

\section*{Abstract}

Computer simulations have become a very powerful tool for scientific research. Given the vast complexity that comes with many open scientific questions, a purely analytical or experimental approach is often not viable. For example, biological systems (such as the human brain) comprise an extremely complex organization and heterogeneous interactions across different spatial and temporal scales. In order to facilitate research on such problems, the BioDynaMo project (\url{https://biodynamo.web.cern.ch/}) aims at a general platform for computer simulations for biological research. Since the scientific investigations require extensive computer resources, this platform should be executable on hybrid cloud computing systems, allowing for the efficient use of state-of-the-art computing technology. This paper describes challenges during the early stages of the software development process. In particular, we describe issues regarding the implementation and the highly interdisciplinary as well as international nature of the collaboration. Moreover, we explain the methodologies, the approach, and the lessons learnt by the team during these first stages.

\section*{Keywords}

Computer simulation, Biological development, Biological dynamics, Brain

\section{Introduction}
Most laboratories in computational biology develop their own custom software to carry out one specific simulation. These applications are monolithic, hard to extend and usually do not scale. Consequently, a lot of resources are spent in developing functionality that has already been created elsewhere. The BioDynaMo project has been started to close the gap between very specialized applications and highly scalable systems to give life scientists access to the rapidly growing computational resources.


Our starting point was the neurodevelopmental simulation software Cortex3D \cite{zubler}. It is a tool that is able to grow sophisticated structures based on simple rules defined by the computational scientist. Our first development stage was dedicated to code modernization. In general, software modernization is a collective term that subsumes a variety of activities. In our case it means transforming the application from Java to C++ and changing the architecture in a way to utilize multiple levels of parallelism offered by today's hardware.

The multidisciplinary nature of this initiative requires expertise from different backgrounds. This poses additional challenges in managing diverse collaborators aligning project members and defining the unique selling propositions in comparison with other simulation packages.

Aiming at a  platform that will be used by a great number of researchers simulating large scale systems, software errors will have a huge impact. Therefore, we give an overview about software verification, different approaches and why it is of great significance for this project. 

\section{Code Modernization}
High performance and high scalability are the prerequisites to address ambitious research questions like modeling epilepsy. Our efforts in code modernization were driven by the goal to remove unnecessary overhead and update the software design to tap the unused potential caused by the paradigm shift to multi and many-core systems. Before 2004, since performance was clock speed driven, buying a processor of the next generation automatically increased application performance without changing a single line of code. Physical limitations forced the CPU vendors to change their strategy to push the edge of the performance envelope \cite{sutter}. Although this paradigm shift helped to improve theoretical throughput, sequential applications benefit only marginally from new processor generations. This puts additional burden on the application developers who have to refactor existing applications and deal with the increased complexity of parallel programs. As core counts are constantly increasing, the portion of unused computing resource will grow in future if these changes are not applied. Furthermore, a modern processor offers multiple levels of parallelism that goes beyond the number of cores: it features multiple threads per core, has multiple execution ports able to execute more than one instruction per clock cycle, is pipelined and provides vector instructions also referred to as SIMD (Single Instruction Multiple Data).

Last years Intel Modern Code Development Challenge was about optimizing sequential C++ brain simulation code provided by the Newcastle University in the UK. The task for participating students was to improve the runtime as much as possible. Using data layout transformations (array of structures AoS to structure of arrays SoA), parallelization with OpenMP, a custom memory allocator and Intel Cilk Plus array notation, the winner was able to improve the runtime by a factor of 320. This reduced the initial execution time on a Intel Xeon Phi Knights Corner Coprocessor with 60 cores from 45 hours down to eight and a half minutes. This clearly shows the economic potential of code modernization efforts. These findings still have to be integrated into the entire code base, since this result has been obtained on a small code sample.

Furthermore, we ported the Java code base to C++. This language is better suited for High Performance Computing as it is compiled to native machine code removing the overhead of running in a virtual machine and provides the right ecosystem for parallelization and optimization. The following iterative porting approach has been chosen. First, a Java class is selected and replaced by its C++ translation. In the second step, this C++ class is connected to the remaining Java application. Finally, the Java/C++ hybrid is compiled and used to execute a number of tests. If all tests pass, the developer can proceed with the next iteration by selecting another Java class. This procedure significantly simplifies debugging in case the test simulations are not consistent with the original Java results. The error has to be within the changes since the last iteration. Although this approach is associated with additional development overhead in connecting classes in C++ to Java, it gives the benefit of obtaining a runnable system after each iteration. Without that additional effort, the first time the C++ version would be able to execute these tests, would be at the very end, after all classes have been ported.

\section{Managing Diverse Collaborators}
A complex scientific simulation software is a collaborative effort between possibly quite disjoint areas of expertise. Consequently, a considerable challenge in BioDynaMo was to arrive a clear list of specifications that could be understood both by neuroscientists and computer scientists. The challenge does not lay only in the communication; we also needed to define an optimal project scope. In such a project the various impacts of each objectives cannot initially be fully understood by all members.

As a single example, the implementation of electrophysiology and certain ephaptic couplings opens for fast interaction over large distances. The spatial scales over which we are looking at such interaction entirely decide the feasibility within a cloud computing environment. If the interactions of interest to neuroscientists are highly non-local it breaks the space partitioning that underlies our parallelization strategy. On the other hand, if the relevant interaction distances of interest are small, this will behave no differently from the other local interactions already allowed by the model. Given that the initial set of specifications simply dictated epaphtic coupling, it took a considerable and non-trivial discussion to isolate a balance between the resulting computational cost and the value of performing a range of simulations depending on such interactions. 

It follows from the scientific application that the challenge in designing specifications is quite different from the corresponding process in consumer applications. In the actual setting, it might be impossible for the developer to reach a full or even satisfactory understanding of the specific needs of the end user within reasonable time.

Our initial attempt at waterfall approach failed and a proper specification was never produced. The deep reliance on a range of domain experts means that our experience primarily centered around knowledge acquisition, similarly to what has previously been described in scientific software development. \cite{Kelly2015}.

In retrospect, our solution was a gradual transition toward an iterative strategy, somewhat similar to greedy optimization. In our approach, we started with an existing prototype specification. Each group of scientific expertise suggested one or several improvements, which was then passed around for consideration and evaluation. Frequent meetings followed up each feature or parameter in order to evaluate what its impact would be in the product value. All changes better than the current prototype, all perspectives taken into consideration, were agreed upon.

The advantage of such an iterative approach already in the planning stage is that allows us to exploit the collective domain knowledge, even if no single person is able to fully grasp the impacts of each aspect of the specification. In all other regards our experience is multidisciplinary collaboration was in line with what has been reported previously in the biomedical software field \cite{Kane2006}.

\section{Aligning Project Members and Defining Uniqueness}
A related but separate issue was to define the software uniqueness in comparison with other available software tools on the market. To maximize the potential of our software we decided on two essentially parallel development strategies. These led to the possibility to use one of two unique attributes to define our simulator.

The generalization of the simulator essentially led to the transformation of a simulation dedicated to neuroscience into a cloud-based simulator of biological tissue and, in the extreme, to any local interaction of objects and spaces.

This led to considerable discussions between project members as to how define our simulator. Are we designing a neurodevelopment simulator extendable to other areas, or are we actually designing a locality-exploiting physics engine on the cloud with a neurodevelopment library included?

In answering the same question on a more abstract level, are we defined by the technology or by the existing user base? The technological interaction is clearly defined, so what it boils down to is how to maximize the utility of our software. In other words, we face a traditional question of marketing.

Marketing activities started out with a low priority in the project. The early discussion centered on the feasibility of the technical specifications, while marketing was neither the expertise nor special interest of any partner in the scientific collaboration, In part this is a consequence of the stakeholder incentives, since network participation in other activities benefits from the academic credit received in a parallel software practice and within the software subfield \cite{Howison2011}.

While participants lack incentives for this, the underdevelopment and importance of market research for open source software in general \cite{Whitmore2009} and scientific software in particular \cite{Howison2015} are well established. We believe the initial efficiency of any supportive activities such as these would be improved by a clear recognition and task allocation early on in the project. Alternatively, we could search for partners with a wider view of the development process in mind, e.g. involve marketing researchers that are also able to benefit from academic credit for these  activities.

\section{Software Verification in Large Scientific Simulations}

Software plays a pervasive role in the modern world and its development requires particular attention to software correctness and resilience. In such technology-dependent world, software verification is becoming a delicate matter, and raising increasing interest from both academia and industry. While, historically, the effort of software verification was mostly concerning safety-critical systems in automotive, transportation and aerospace industries -- and in general where human lives are at stake -- recently attention moved to more traditional, lower stake and off-the-shelf commercial (or not) software.

Computer simulations are not exempt from risks either, although the concept of \textit{catastrophic} has not to be seen in terms of direct, immediate life threats. Simulations can run on a single or multiple machines, or on the cloud, and run for a few minutes only or for hours or days. The computational costs, and the financial costs as a consequence can be considerable implying a significant loss in case of errors and need to re-run all or part of the simulation. In particular, small errors can propagate and  accumulate into substantial error later in the simulation. Techniques that ensure software quality seem therefore necessary.

\section*{Need for Verification Tools}
There is a common belief in industry that developing software with high level of assurance is too expensive, therefore not acceptable, especially for non safety-critical or financially-critical applications.  Tools and techniques for the formal development of software have played a key role on demystifying this belief. Complex systems from any domain have a behavior that is difficult to predict and techniques for ensuring correctness can be expensive and laborious; however, fixing bugs after deployment would be even more expensive. Post-deployment faults may even lead to disasters; the literature contains many examples of avoidable faults that had catastrophic consequences. Complex systems also show high level of concurrency, i.e. multiple intertwined threads of executions, often running on different hardware, which need to be synchronized and coordinated, and which must share information, often through different paradigms of communication, and race condition are often subtle and critical sources of bugs that testers may easily miss. Here is where verification tools are of paramount importance to ensure software quality.

\section*{Overview of Major Verification Approaches}. 
Tools for software verification allow the application of theoretical principles in practice, in order to ensure that nothing bad will ever happen (safety). The extra effort required by the use of these tools is certainly not for free and comes with increased development costs \cite{Meyer2009}. There are several approaches described in literature, and the list here cannot be exhaustive; for instance \textit{abstract interpretation} \cite{Cousot1977} and \textit{model checking} \cite {Clark1999}, that seek the automation to formally proving certain conditions of systems. However, these techniques tend to verify simple properties only. On the other end of the spectrum, there are interactive techniques for verification such \textit{theorem provers} \cite{Loveland:1978}. These techniques aim at more complex properties, but demand the interaction of users to help the verification. Approaches aiming at finding a good trade-off between these techniques exist too, e.g. auto-active: users are not needed during the verification process (it is automatically performed); they are required instead to provide guidance to the proof using annotations \cite{Khazeev2016}.

\section*{Towards a Verification Approach to BioDynaMo}
A rigorous development approach it is also necessary for the BioDynaMo project if we want to achieve high levels of resilience. Some of the approches have to be ruled out by the very nature of the project and its own peculiarities, for example the programming languages and the fact that it is born as a code modernizaton experience, therefore the software lifecyle followed by the artifact is also non-standard. In absence of a total control on the process and the technologies in use, ex-post model checking appears here to be the most viable solution.

\section{Discussion}
The field of computational biology covers a wide range of scientific topics, each producing many different scientific models, such as for instance described by \cite{bauer2014developmental}, \cite{freund2014numerical} and \cite{izhikevichEdelman2008large}. Hence, a general platform for biological research should be able to meet a number of different requirements. It is crucial that this diversity of the prospective users is already taken into account during the software development process. Incorporating such diversity means that different the multidisciplinary project team of BioDynaMo must be able to efficiently interact, and make decisions based on the expertise of each team member.

In addition to these more scientifically-centered aspects, also considerable challenges arise from a computational/technological point of view. First steps towards such efficient software implementation have been made in the context of the ``Intel Modern Code Developer Challenge'' competition. Overall, we believe we have created a collaborative foundation for the efficient continuation of the very ambitious software development project of BioDynaMo.

However, considerable challenges remain to be approached and tackled in the current software development process. The verification and validation of the software is paramount. The recent study of \cite{Eklundetal2016cluster} demonstrates the extraordinary risks that arise when the correctness and validity of software tools for scientific research are not properly assessed. We have identified this key aspect to require further efforts in parallel to the overall development process.

\section*{Acknowledgements}

M.Mazzara and L.Johard received logistic and financial support by Innopolis University, in particular by the Service Science and Engineering lab (SSE) within the Institute of Technologies and Software Development.
\newline
R.Bauer and M.Kaiser were supported by the Human Green Brain Project \newline(\url{www.greenbrainproject.org}) through the Engineering and Physical Sciences Research Council (EP/K026992/1). The funders had no role in study design, data collection and analysis, decision to publish, or preparation of the manuscript.
\newline
M.Manca contribution has been supported by SCImPULSE Foundation, a public benefit organization based in The Netherlands.
\newline
A.Di Meglio, L.Breitwieser, and M.Manca were supported by CERN, and CERN openlab under the program "Code Modernization", made possible by a cooperation with Intel, to which we extend our thanks.

\clearpage
\markboth{}{}

\section*{Competing interests}

“The authors declare that they have no competing interests.”

\end{document}